\documentclass{article}

\usepackage{natbib} 

\usepackage[final]{nips_2018}

\usepackage[utf8]{inputenc} 
\usepackage[T1]{fontenc}    
\usepackage{hyperref}       
\usepackage{url}            
\usepackage{booktabs}       
\usepackage{multirow}

\usepackage{amsfonts}       
\usepackage{nicefrac}       
\usepackage{microtype}      
\usepackage{color}

\usepackage{graphicx}
\usepackage{subcaption} 
\usepackage[bottom, perpage]{footmisc}  

\title{Logit Pairing Methods Can Fool \\ Gradient-Based Attacks}

\author{
  Marius Mosbach\thanks{Both authors contributed equally to this work.}$^{\ast \dagger}$, Maksym Andriushchenko\footnotemark[1]$^{\ast \ddagger}$, Thomas Trost$^{\dagger}$, \\ \textbf{Matthias Hein$^{\sharp}$}, \textbf{Dietrich Klakow}$^{\dagger}$\\
  $\dagger$ Spoken Language Systems (LSV), Saarland University, Saarland Informatics Campus\\ 
  $\ddagger$ Dept. of Mathematics and Computer Science, Saarland University, Saarland Informatics Campus\\
  $\sharp$ Dept. of Computer Science, University of Tübingen
}

\begin{document}
\bibliographystyle{plainnat}

\maketitle

\begin{abstract}
	
	Recently, \citet{kannan2018adversarial} proposed several logit regularization methods to improve the adversarial robustness of classifiers. We show that the computationally fast methods they propose -- Clean Logit Pairing (CLP) and Logit Squeezing (LSQ)  --  just make the gradient-based optimization problem of crafting adversarial examples harder without providing actual robustness. We find that Adversarial Logit Pairing (ALP) may indeed provide robustness against adversarial examples, especially when combined with adversarial training, and we examine it in a variety of settings. However, the increase in adversarial accuracy is much smaller than previously claimed. Finally, our results suggest that the evaluation against an iterative PGD attack relies heavily on the parameters used and may result in false conclusions regarding robustness of a model. 

\end{abstract}

\section{Introduction}

\citet{szegedy2013intriguing} showed that state-of-the-art image classifiers are not robust against certain small perturbations of the inputs, known as \textit{adversarial examples}. Since then, many new attacks have been proposed aiming at better ways of crafting adversarial examples, and also many new defenses to increase the robustness of classifiers. Notably, almost all defenses previously proposed have been broken by applying different attacks \citep{carlini2017towards, athalye2017synthesizing, athalye2018robustness, athalye2018obfuscated}. 

\begin{figure}[hb] 
	\centering
	\resizebox{.95\columnwidth}{!}{%
		\centering
		\hspace{-0.5cm}
		\begin{subfigure}[t]{0.225\textwidth}		
			\centering
			\includegraphics[width=1.15\textwidth]{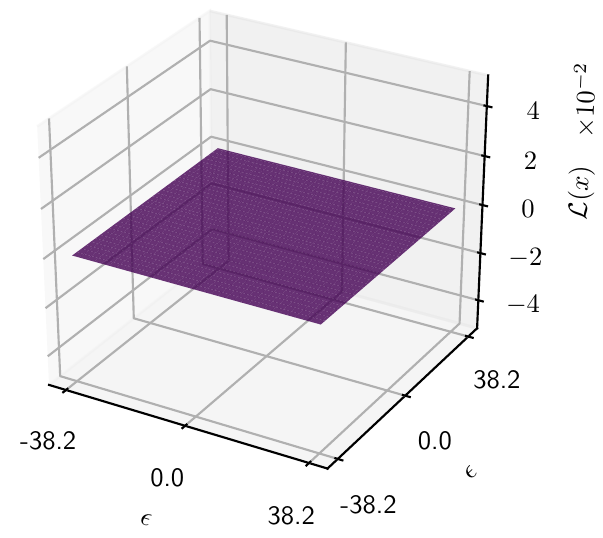}
			\caption{\centering Adv. Training}
		\end{subfigure}%
		\qquad ~
		\begin{subfigure}[t]{0.225\textwidth}		
			\centering
			\includegraphics[width=1.2\textwidth]{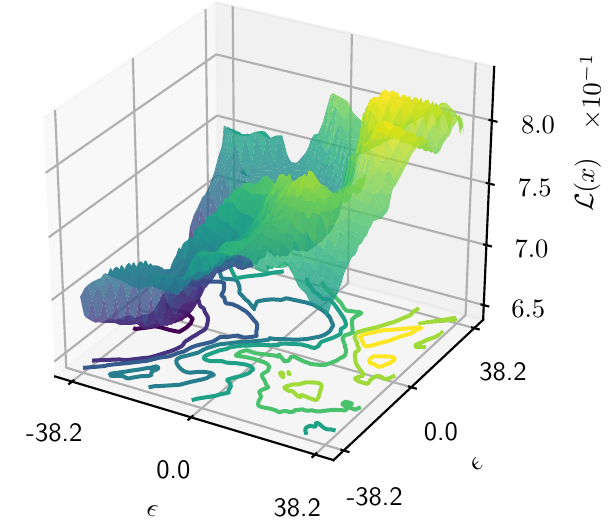}
			\caption{\centering Logit Squeezing}
		\end{subfigure}%
		\qquad ~
		\begin{subfigure}[t]{0.225\textwidth}		
			\centering
			\includegraphics[width=1.2\textwidth]{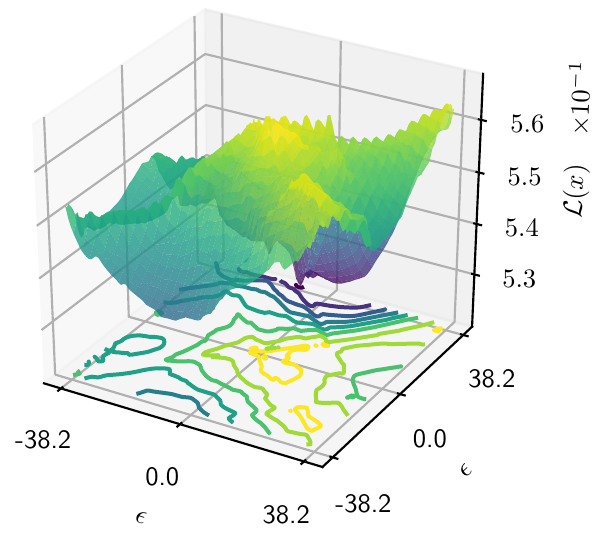}
			\caption{\centering Clean Logit Pairing}
		\end{subfigure}%
		\qquad ~
		\begin{subfigure}[t]{0.225\textwidth}		
			\centering
			\includegraphics[width=1.15\textwidth]{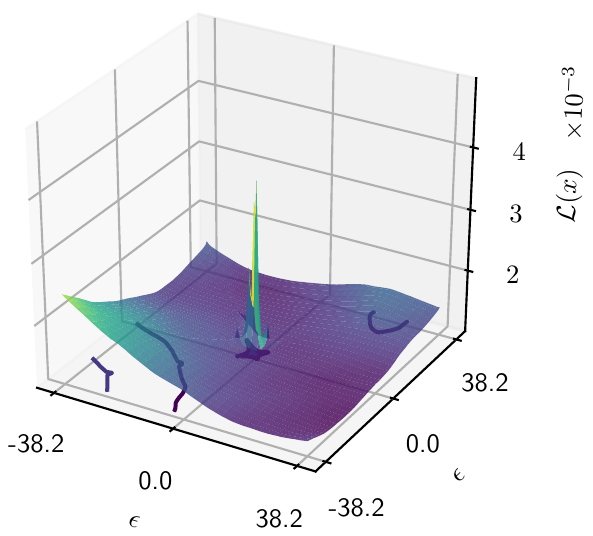}
			\caption{\centering Adv. Logit Pairing}
		\end{subfigure}%
	}
	
	\caption{Input loss surfaces of MNIST models in a random subspace around an input image with $\epsilon=38.25$. We can clearly see a distorted loss surface for the logit regularization methods, which can prevent gradient-based attacks from succeeding. Additional visualizations are found in Figures \ref{fig:mnist_clp_loss_surfaces_appendix}, \ref{fig:mnist_alp_loss_surfaces_appendix}, \ref{fig:cifar10_clp_loss_surfaces_appendix}, and \ref{fig:cifar10_alp_loss_surfaces_appendix} in the Appendix.}
	\label{fig:loss_surfaces_page1}
\end{figure}

A prominent defense that could not be broken so far is adversarial training \citep{goodfellow2014explaining, madry2017towards}.
 There is also a line of work on the provable robustness of classifiers \citep{hein2017formal, kolter2017provable, raghunathan2018certified}, classifiers which by definition cannot be broken because they derive and report a lower bound on the \textit{worst-case adversarial accuracy}, i.e. the accuracy of the model on the strongest possible adversarial examples under the given threat model. On the other hand, any attack provides only an upper bound on the worst-case adversarial accuracy, which we will refer to as just \textit{adversarial accuracy}. The problem with many recently proposed defenses is that they evaluate \textit{only} upper bounds, which might be arbitrarily loose, i.e. there might exist an attack which is able to reduce the adversarial accuracy significantly compared to some baseline attack. 
However, one of the main issues with lower bounds is that they are usually too small to be useful, so a special way of maximizing them is applied during training. This may interfere with a proposed defense which one aims to evaluate. Thus, providing non-trivial lower bounds --~with or without special training --~is an important and active area of research \citep{wong2018scaling, zhangefficient, xiao2018training, croce2018provable}. Unfortunately, these methods do not yet scale to large-scale datasets like ImageNet, and thus one still has to rely solely on upper bounds on adversarial accuracy to estimate the robustness on these datasets.

Given the recent history of breaking most of the defenses accepted at ICLR 2018 \citep{athalye2018obfuscated, uesato2018adversarial}, it is now natural to question any new non-certified defense. Many recent papers \citep{buckman2018thermometer, kannan2018adversarial, yao2018hessian} that claim robustness of their models mainly rely on the PGD attack from \citet{madry2017towards} with the default settings. They assume that they evaluate their models against a ``strong adversary'' and that the adversarial accuracy they obtain is close to the minimal possible. In this paper, we show that it is not the case for CLP, LSQ and some ALP models proposed by \citet{kannan2018adversarial}.

\paragraph{Threat model and attack settings.}

We consider the classification of images with pixel values in $[0, 255]$ and focus on the white-box threat model, i.e. an attacker has complete knowledge of the model. We consider adversarial perturbations bounded with respect to an  $L_{\infty}$ norm of $\epsilon=76.5$ for MNIST, and  $\epsilon=16$ for CIFAR-10 and Tiny ImageNet. We follow the settings of \citet{kannan2018adversarial} and evaluate MNIST and CIFAR-10 models with untargeted attacks. Tiny ImageNet models were evaluated with targeted attacks which is consistent with \citet{athalye2018obfuscated}. For crafting adversarial examples, we used the PGD attack \citep{madry2017towards} with maximum adversarial perturbation $\epsilon$ and experimented with different number of iterations $n$, step sizes $\epsilon_i$, and restarts $r$. For further comparison we also evaluate all MNIST and CIFAR-10 models against the SPSA attack \citep{uesato2018adversarial}. When crafting untargeted adversarial examples, we maximize the loss using the true label.

\begin{figure}[b] 
	\centering
	\resizebox{.95\columnwidth}{!}{%
		\centering
		\hspace{-1.0cm}
		\begin{subfigure}[t]{0.33\textwidth}		
			\centering
			\includegraphics[width=1.2\textwidth]{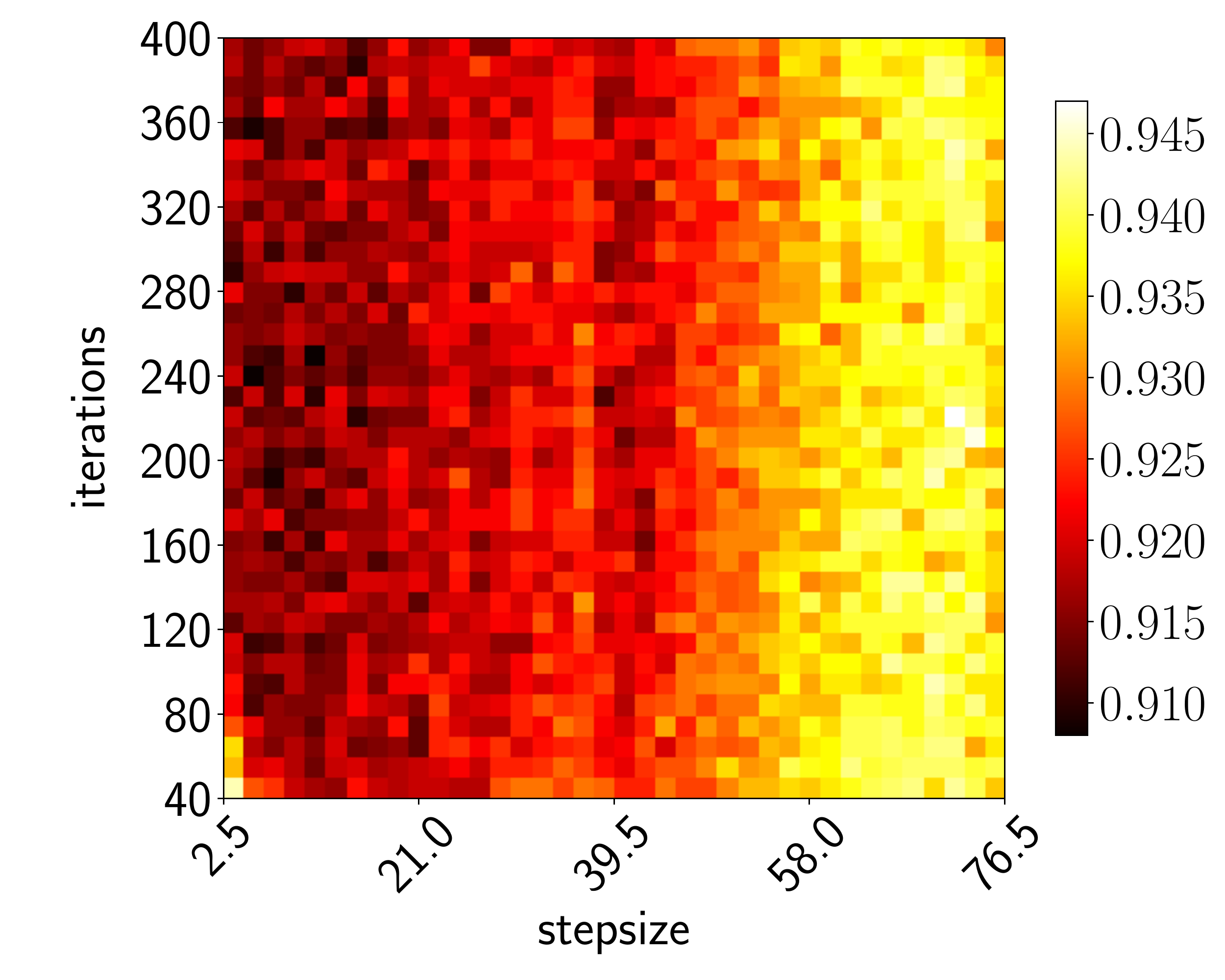}
			\caption{\centering Adv. Training}
		\end{subfigure}
		\qquad  ~
		\begin{subfigure}[t]{0.33\textwidth}		
			\centering
			\includegraphics[width=1.2\textwidth]{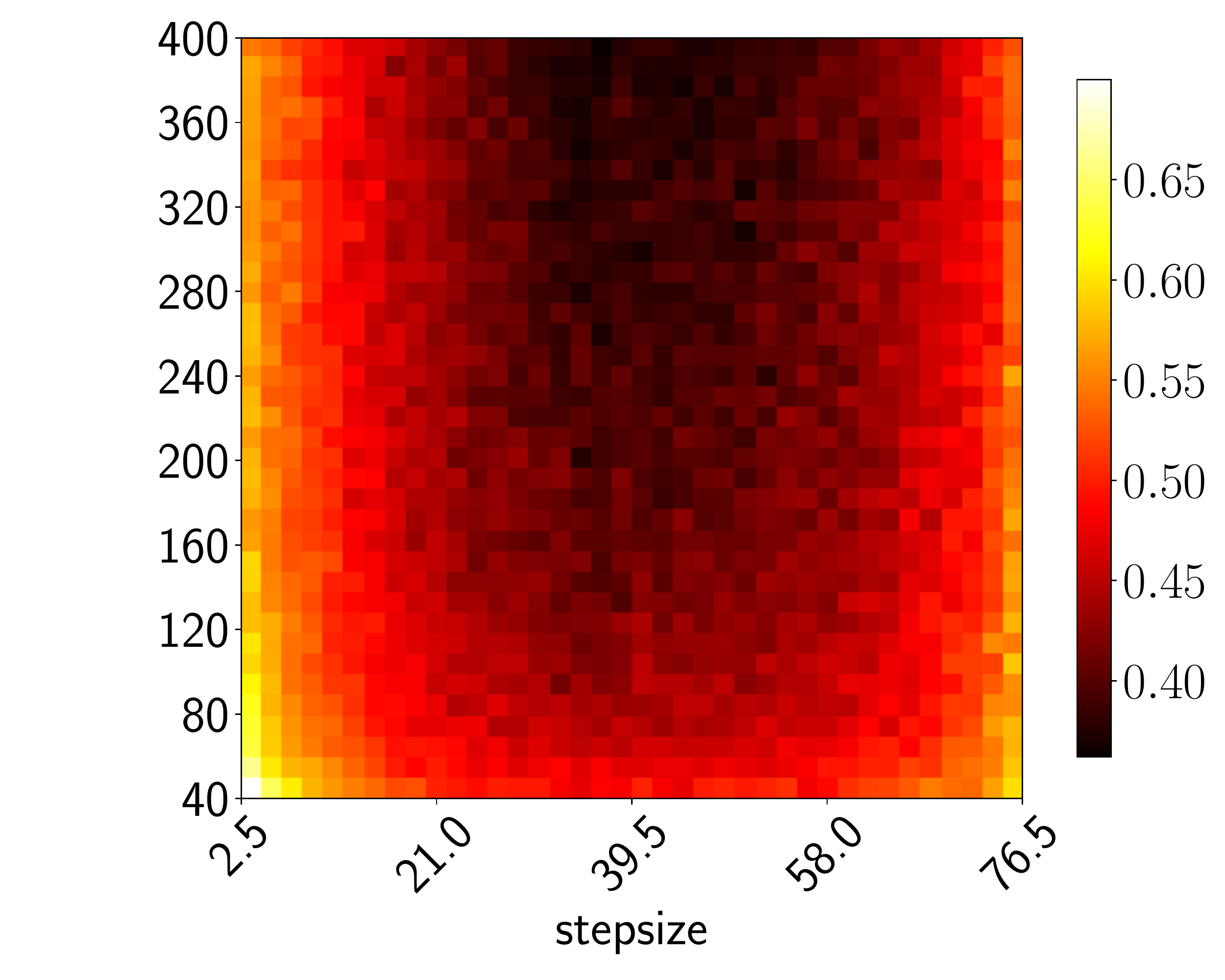}
			\caption{\centering Logit Squeezing}
		\end{subfigure}%
		\qquad  ~
		\begin{subfigure}[t]{0.33\textwidth}		
			\centering
			\includegraphics[width=1.2\textwidth]{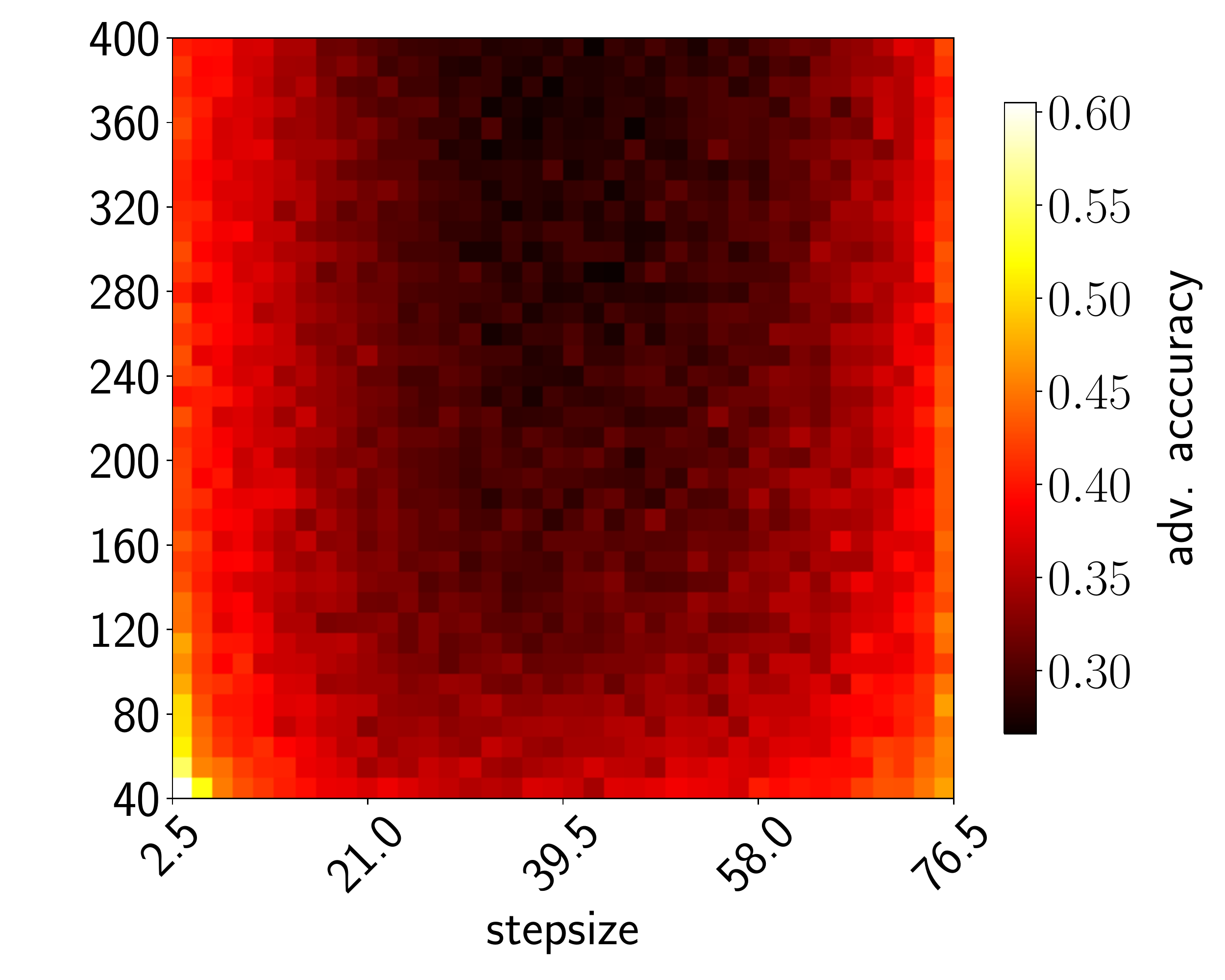}
			\caption{\centering Clean Logit Pairing}
		\end{subfigure}%
		
	}
	\caption{Heatmaps of the adversarial accuracy for 100\% adversarial training \citep{madry2017towards},~LSQ, and CLP models trained on MNIST for different settings of step size $\epsilon_i$ and number of iterations $n$ when running the PGD attack with $\epsilon = 76.5$. Heatmaps for other models can be found in Figures \ref{fig:grid_search_at} and \ref{fig:grid_search_alp} in the Appendix. For all heatmaps, the adversarial accuracy was evaluated on 1000 points drawn randomly from the test data.}
	\label{fig:grid_search_clp}
\end{figure}

\begin{figure}[h] 
	\centering
	\resizebox{.85\columnwidth}{!}{%
		\centering
		\hspace{-1.0cm}
				\begin{subfigure}[t]{0.45\textwidth}		
					\centering
					\includegraphics[width=1.2\textwidth]{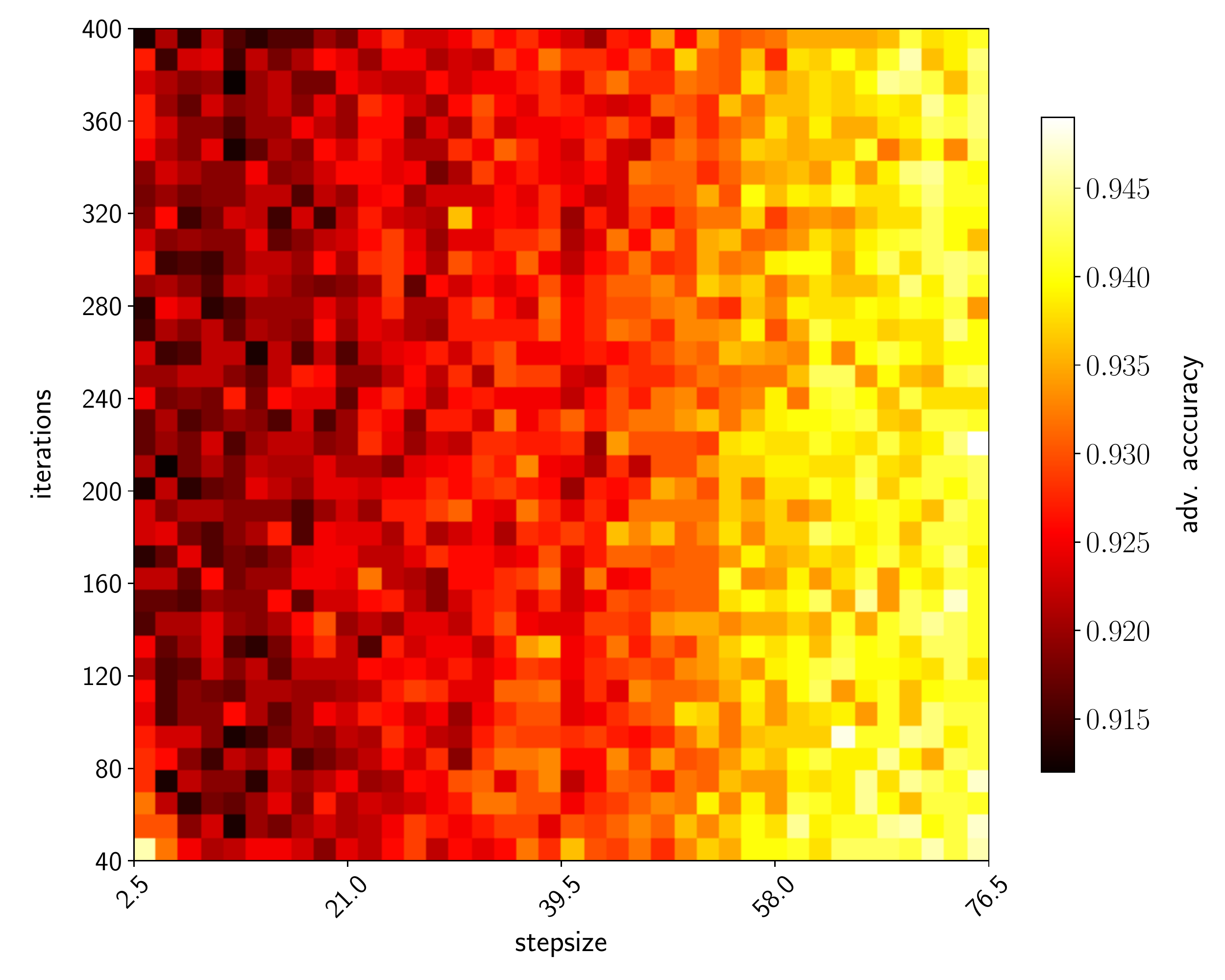}
					\caption{MNIST 50\% AT model}
				\end{subfigure}
				\qquad \qquad ~
		\begin{subfigure}[t]{0.45\textwidth}		
			\centering
			\includegraphics[width=1.2\textwidth]{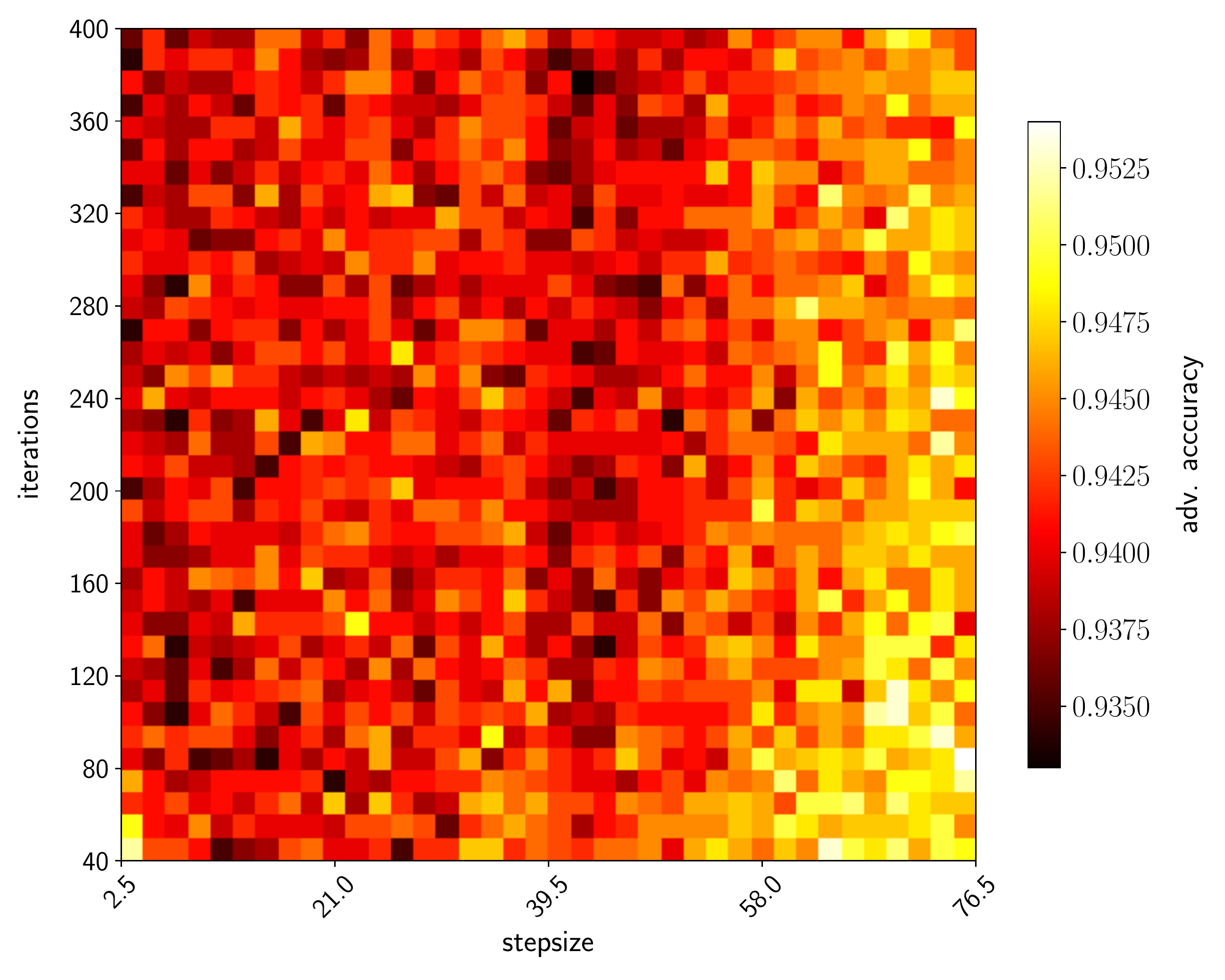}
			\caption{MNIST clean + ALP model}
		\end{subfigure}%
		
	}
	\caption{Heatmaps of the adversarial accuracy for 50\% AT and clean + ALP models trained on MNIST for different settings of step size $\epsilon_i$ and number of iterations $n$ when running PGD with $\epsilon = 76.5$.}
	\label{fig:grid_search_at}
\end{figure}

\begin{figure}[h] 
	\centering
	\resizebox{.85\columnwidth}{!}{%
		\centering
		\hspace{-1.0cm}
		\begin{subfigure}[t]{0.45\textwidth}		
			\centering
			\includegraphics[width=1.2\textwidth]{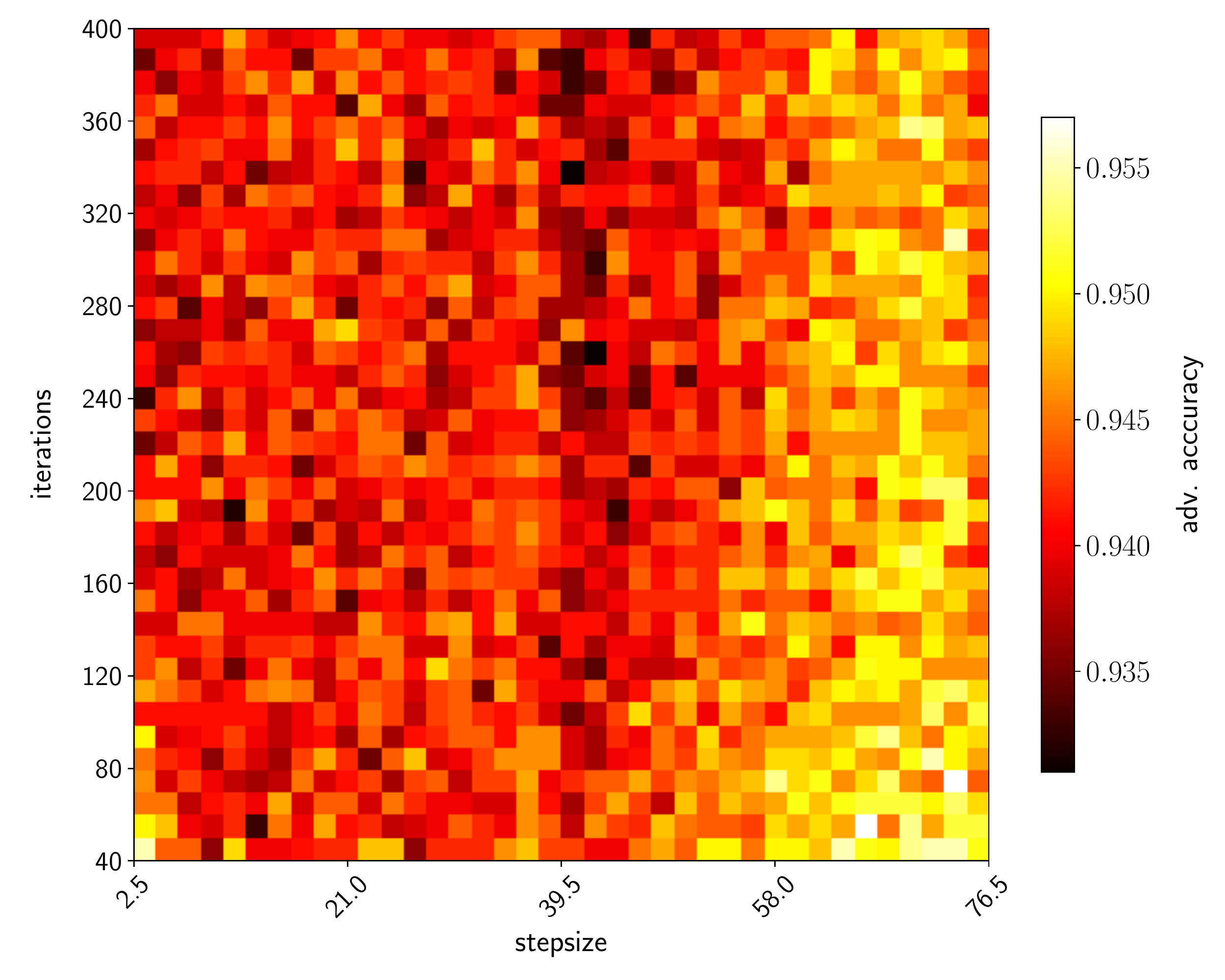}
			\caption{MNIST 100\% AT + ALP model}
		\end{subfigure}
		\qquad \qquad ~
		\begin{subfigure}[t]{0.45\textwidth}		
			\centering
			\includegraphics[width=1.2\textwidth]{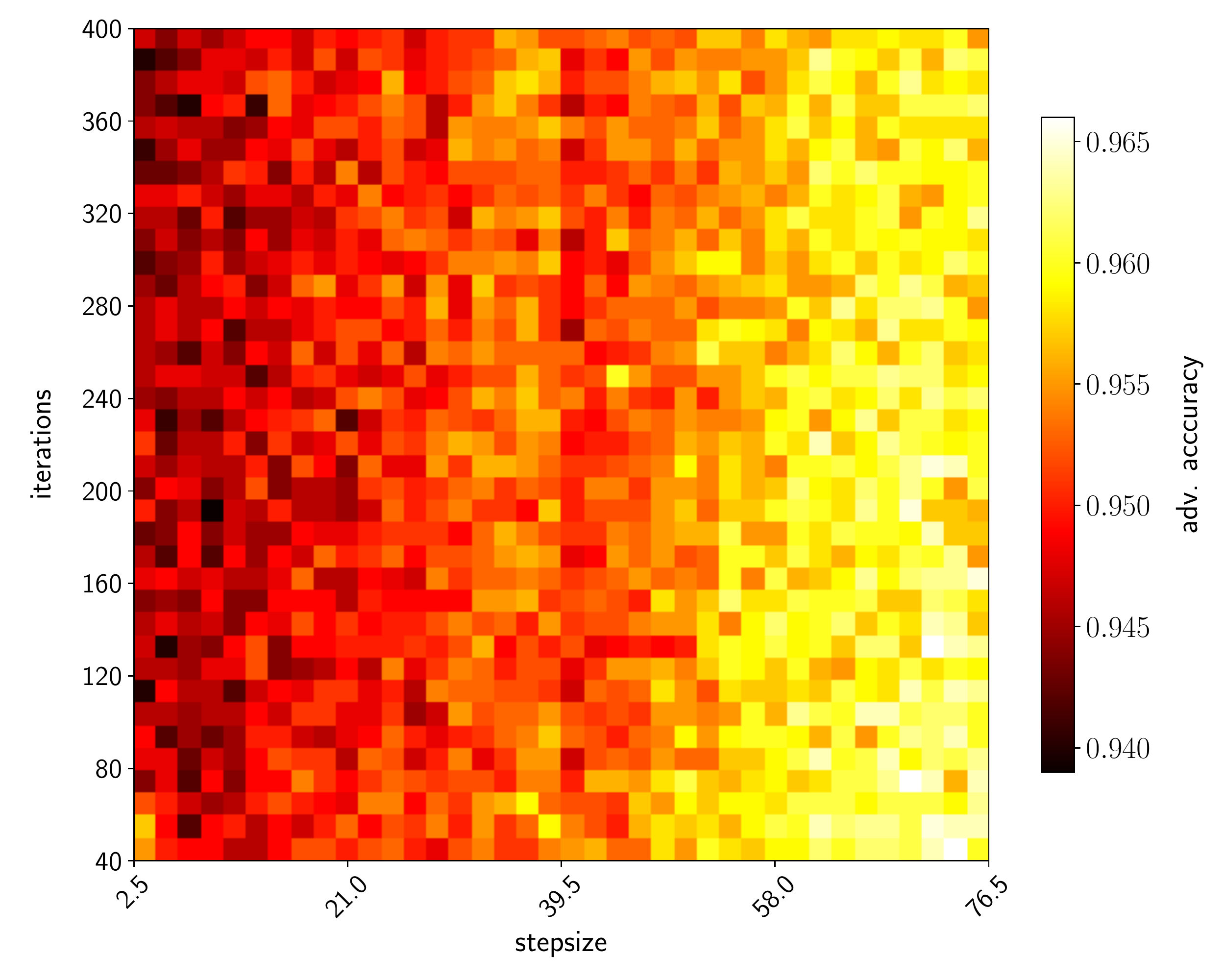}
			\caption{MNIST 50\% AT + ALP model}
		\end{subfigure}%
		
	}
	\caption{Heatmaps of the adversarial accuracy for 100\% AT + ALP and 50\% AT + ALP models trained on MNIST for different settings of step size $\epsilon_i$ and number of iterations $n$ when running PGD with $\epsilon = 76.5$.}
	\label{fig:grid_search_alp}
\end{figure}

\paragraph{Contribution.}
First, we give empirical evidence that CLP, LSQ, and some ALP models distort the loss surface in the input space and thus fool gradient-based attacks without providing actual robustness. This can be seen as a particular case of masked or obfuscated gradients \citep{papernot2017practical, athalye2018obfuscated}. We illustrate this by analyzing the input space loss surface in two random directions (Figure~\ref{fig:loss_surfaces_page1}). We provide an extensive experimental evaluation of the robustness of CLP, LSQ, and ALP models on MNIST, CIFAR-10, and Tiny ImageNet datasets against the PGD attack with a large number of iterations and random restarts reducing the adversarial accuracy of e.g. MNIST-LSQ model from 70.6\% to 5.0\% (Table~\ref{table:mnist}). Our results suggest that while CLP and LSQ do not provide actual robustness, ALP may provide additional robustness on top of adversarial training. However, the increase is much smaller than claimed by \citet{kannan2018adversarial} (Table \ref{table:tiny_imagenet}), and it remains unclear whether this is actual robustness or the increase in adversarial accuracy is only due to the distortion of the loss surface in the input space. Finally, we highlight the importance of performing many random restarts and an exhaustive grid search over the attack parameters of PGD, especially when the loss surface is distorted. We illustrate this by plotting the distribution of the loss values over different restarts of PGD~(Figure~\ref{fig:histograms}) as well as heatmaps of the adversarial accuracy for different PGD attack parameters (Figures \ref{fig:grid_search_clp}, \ref{fig:grid_search_at}, \ref{fig:grid_search_alp}).

\paragraph{Related work.}
We note that a regularization term similar to ALP was proposed before in \citet{heinze2017conditional} (the conditional variance penalty) in the context of adversarial domain shifts. However, they do not study its effect in the context of $L_p$-bounded adversarial examples, thus we only analyze the results of ALP from \citet{kannan2018adversarial}.

Recently, \citet{engstrom2018evaluating} evaluated the robustness of ALP on a single ImageNet model. However, there are important differences compared to our work. First, they do not explore the robustness of the computationally cheap methods CLP and LSQ, which are a significant contribution of \citet{kannan2018adversarial}. Second, they only test an ALP model that was trained on clean examples, while \citet{kannan2018adversarial} mainly advocate for the usage of ALP combined with mixed-minibatch PGD (i.e. training on 50\% clean and 50\% adversarial examples), which we explore in detail. Finally, while \citet{engstrom2018evaluating} only consider a single ImageNet model, we perform experiments on MNIST, CIFAR10, and Tiny ImageNet and show that the conclusions depend on the dataset,  highlighting the importance of evaluating multiple models on
multiple datasets before attempting to draw general conclusions.

\begin{table}[b]
	\centering
		\begin{tabular}{lrrrrr}
			\toprule
			
			& & \multicolumn{ 4}{c}{\textbf{Adversarial accuracy}~ ($L_{\infty}$, $\epsilon = 76.5$)} \\
			
			\cmidrule(r){3-6}
			
			\multirow{ 3}{*}{\textbf{Model}} & \multirow{ 3}{*}{\textbf{Accuracy}} & \multicolumn{1}{c}{\textbf{SPSA attack}} & \multicolumn{3}{c}{\textbf{PGD attack}} \\
			\cmidrule(r){3-3}
			\cmidrule(r){4-6}
			
			& & \quad $bs = 8192$ & \quad $\epsilon_{i} = 2.55$ & \quad $\epsilon_{i} = 50.0$ & \quad $\epsilon_{i} = 50.0$  \\
			& & \quad $n=200$ & $n=40$ & $n=200$ & $n=200$ \\
			& & $r=1$ & $r=1$ & $r=1$ & $r=10000$  \\
			
			\midrule

			Plain & \textbf{99.2\%}  & 0.0\% &  0.0\% & 0.0\%  & 0.0\% \\
			
			CLP & 98.8\%  & 44.8\% & 62.4\% & 29.1\% & 4.1\%  \\
			
			LSQ & 98.8\%  & 54.1\% & 70.6\% & 39.0\% & 5.0\%  \\
			
			\midrule
			
			Plain + ALP & 98.5\%  & 94.4\% & 96.0\% & 93.8\% & 88.9\% \\
			
			50\% AT + ALP & 98.3\% & 95.1\% & \textbf{97.2\%} & \textbf{95.3\%} & \textbf{89.9\%}  \\
			
			50\% AT & 99.1\% & 92.7\% & 95.6\% & 93.1\% & 88.2\%  \\
			
			100\% AT + ALP & 98.4\% & \textbf{95.7\%} & 96.6\% & 93.8\% & 85.7\%  \\
			
			100\% AT & 98.9\% & 92.9\% & 95.2\% & 92.7\% & 88.0\%  \\
			\bottomrule
		\end{tabular}
	\vspace{0.1cm}
	\caption{Clean and adversarial accuracy against different attacks on MNIST. Adversarial accuracy is evaluated against the PGD attack using $\epsilon = 76.5$ with different number of iterations $n$, step sizes  $\epsilon_i$, and restarts $r$. AT denotes adversarial training. CLP and LSQ models are trained with $\lambda=0.5$ and with $\mathcal{N}(0, 0.5)$ noise augmentation. All ALP models were trained with $\lambda=1.0$.}
	\label{table:mnist}
\end{table}

\section{Experiments}

We note that none of CLP nor LSQ models were officially released by \citet{kannan2018adversarial}. Thus, following \citet{kannan2018adversarial} we train our CLP and LSQ models from scratch with Gaussian data augmentation (denoted by $\mathcal{N}(\mu, \sigma)$ in all tables). On MNIST we use the same LeNet architecture as \citet{kannan2018adversarial} and train all models for 500 epochs with a batch size of $200$. For CIFAR-10, we use the ResNet20-v2 architecture \citep{DBLP:conf/eccv/HeZRS16} and all models are trained for $100$ epochs with a batch size of $128$. On Tiny ImageNet we use the same ResNet50-v2 architecture as \citet{kannan2018adversarial} and analyze their pre-trained ALP models as well as our own models that we trained from scratch with standard data augmentation and weight decay for 100 epochs using a batch size of 256. Note that we do not perform any fine-tuning from pre-trained ImageNet models to better understand the contribution of the logit pairing methods. For all models, we use the Adam optimizer \citep{kinga2015method}, and evaluate on 1000 images drawn randomly from the test data.  For crafting adversarial examples, we use the PGD and SPSA attacks implemented in the Cleverhans library~\citep{papernot2018cleverhans}. We apply adversarial training using the PGD attack with the step size of $2.55$ and $40$ iterations for MNIST, and the step size of $2.0$ and $10$ iterations for CIFAR-10 and Tiny ImageNet.

\begin{figure}[h] 
		\centering
		\resizebox{.95\columnwidth}{!}{%
		\centering
		\includegraphics[width=0.9\textwidth]{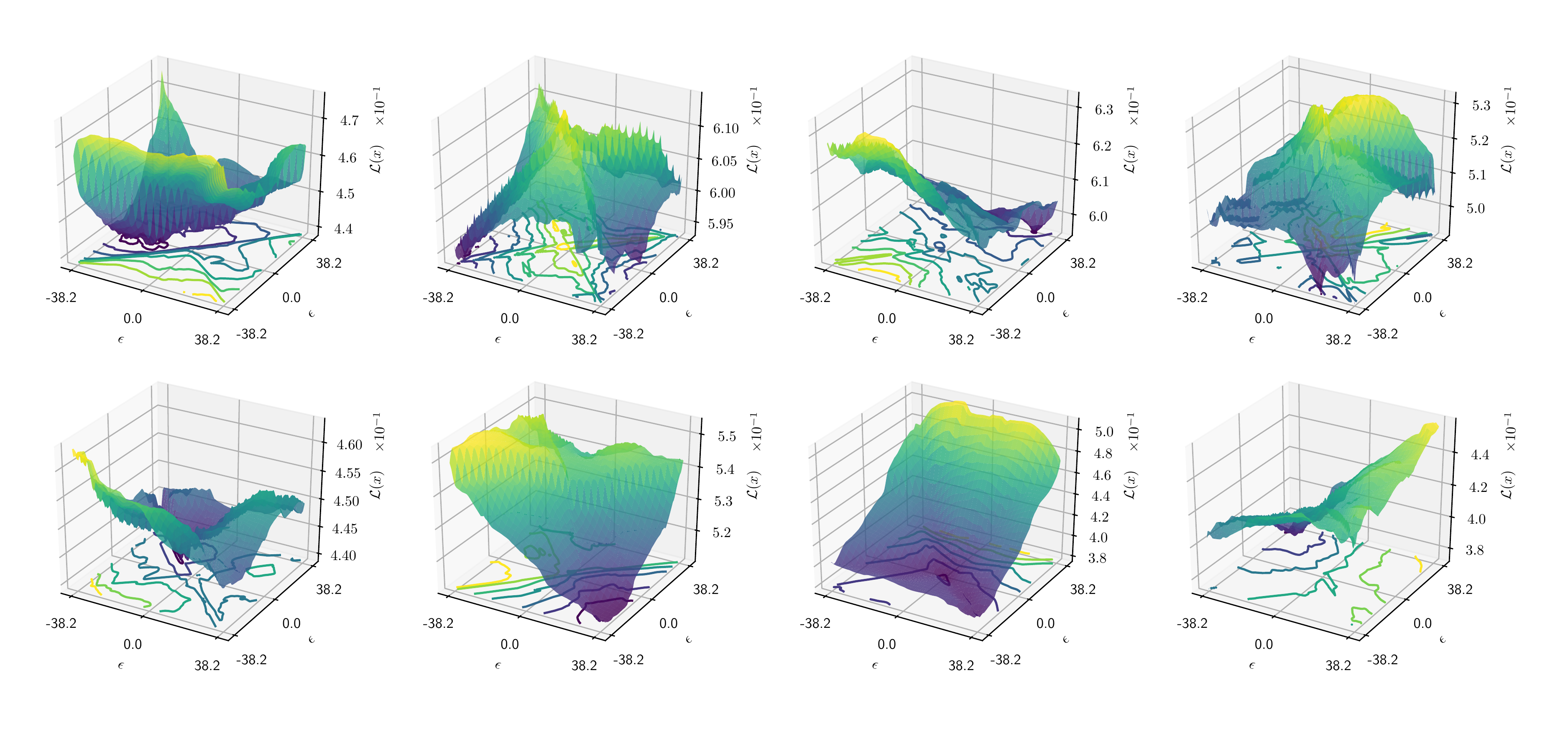}%
		}
		\caption{Input loss surfaces of CLP model on MNIST in a random subspace with $\epsilon=38.25$ for the first eight test examples. The loss surface contains many local maxima and hence makes gradient-based attacks much more difficult. This is in line with our quantitative results in Table \ref{table:mnist}, showing that this model does not provide actual robustness and that the gradient-based PGD attack must use many random restarts to successfully craft adversarial examples.}
		\label{fig:mnist_clp_loss_surfaces_appendix}
	\end{figure}

\begin{figure}[h] 
		\centering
		\resizebox{.95\columnwidth}{!}{%
		\centering
		\includegraphics[width=0.9\textwidth]{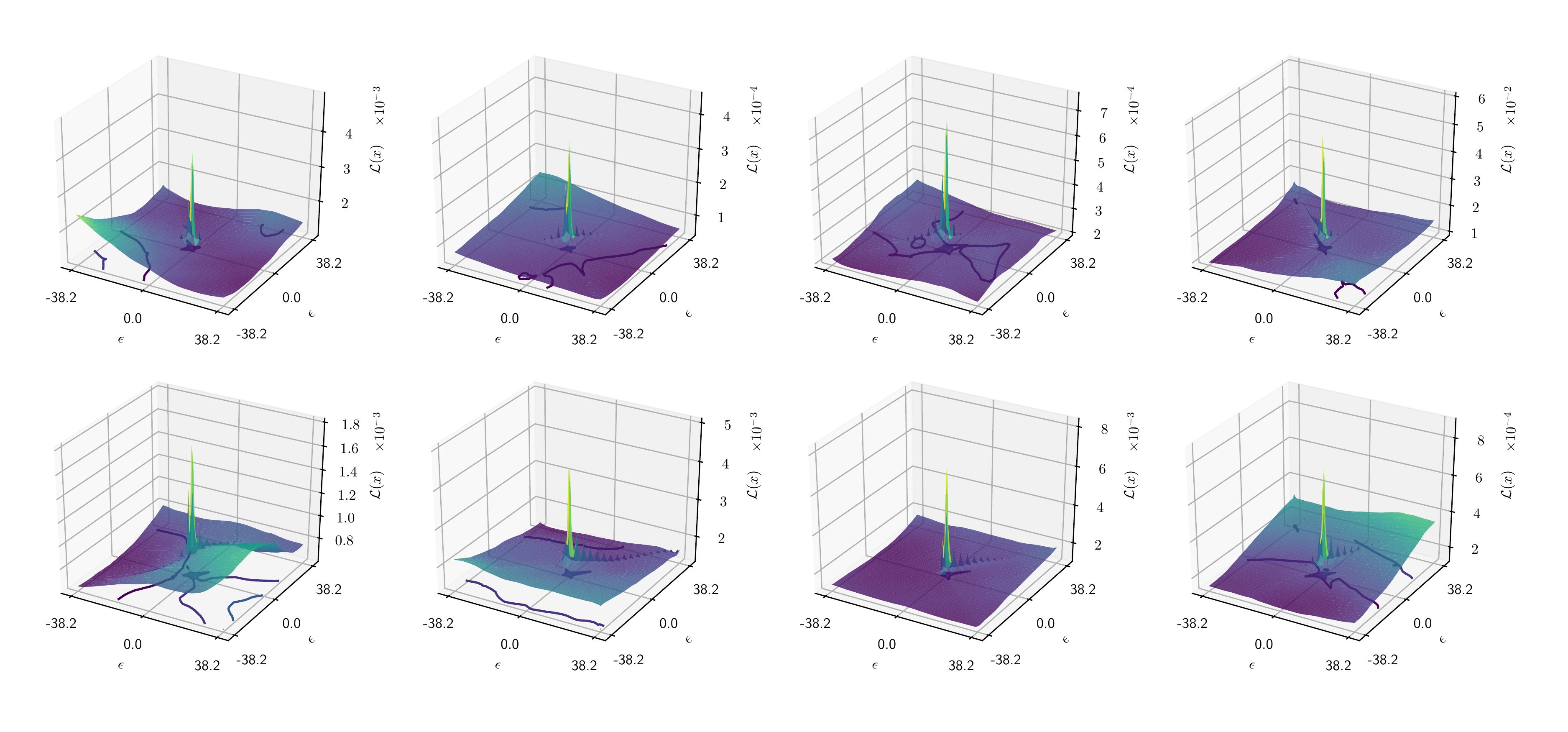}%
		}
		\caption{Input loss surfaces of Plain + ALP model  on MNIST in a random subspace with $\epsilon=38.25$ for the first eight test examples. The loss surface has a local maximum at the input point. At the same time, our quantitative results in Table \ref{table:mnist} show that this model is resistant even to our strongest attack with many random restarts of PGD attack.}
		\label{fig:mnist_alp_loss_surfaces_appendix}
	\end{figure}

We visualize the cross-entropy loss in a two-dimensional subspace of the input space in the vicinity of an input point $x$, where the subspace is spanned by two random signed vectors scaled by $\epsilon=38.25$ for MNIST and $\epsilon=16.0$ for CIFAR-10.
We observe that the loss surfaces of models trained with CLP, LSQ, and ALP can contain many local maxima which makes gradient-based attacks such as PGD and SPSA difficult (Figure \ref{fig:loss_surfaces_page1} and Figures \ref{fig:mnist_clp_loss_surfaces_appendix}, \ref{fig:mnist_alp_loss_surfaces_appendix}, \ref{fig:cifar10_clp_loss_surfaces_appendix}, \ref{fig:cifar10_alp_loss_surfaces_appendix} in the Appendix). In order to deal with this and in contrast to the results given by \citet{kannan2018adversarial} and \citet{engstrom2018evaluating}, we first perform a grid search over the step size $\epsilon_i$ and the number of iterations $n$ of the PGD attack. We then run our attacks with multiple random restarts $r$ and report the adversarial accuracy over the most harmful restarts. We illustrate the importance of performing a grid search over the PGD attack parameters in Figure \ref{fig:grid_search_clp}. Additionally, we show the importance of having many random restarts by plotting the distribution of the loss of the PGD attack across many restarts in Figure \ref{fig:histograms}, which highlights that there are cases where many random restarts are needed in order to find an adversarial example.

We make our code and models publicly available\footnote{\url{https://github.com/uds-lsv/evaluating-logit-pairing-methods}}.

\begin{table}[b]
	\centering
		\begin{tabular}{lrrrrr}
			\toprule
			
			& & \multicolumn{ 4}{c}{\textbf{Adversarial accuracy}~ ($L_{\infty}$, $\epsilon = 16.0$)} \\
			
			\cmidrule(r){3-6}
			
			\multirow{ 3}{*}{\textbf{Model}} & \multirow{ 3}{*}{\textbf{Accuracy}} & \multicolumn{1}{c}{\textbf{SPSA attack}} & \multicolumn{3}{c}{\textbf{PGD attack}} \\
			\cmidrule(r){3-3}
			\cmidrule(r){4-6}
			
			& & \quad $bs = 2048$ & \quad $\epsilon_{i} = 2.0$ & \quad $\epsilon_{i} = 4.0$ & \quad $\epsilon_{i} = 4.0$ \\ 
			& & $n = 100$ & $n=10$ & $n=400$  & $n=400$  \\
			& & $r=1$ & $r=1$ & $r=1$ & $r=100$ \\
			
			\midrule
			
			Plain & \textbf{83.0\%}  & 0.0\% &  0.0\% & 0.0\%  & 0.0\% \\
			
			CLP & 73.9\%  & 0.3\% & 2.8\% & 0.4\% & 0.0\% \\
			
			LSQ & 81.7\%  & 13.6\% & \textbf{27.0\%} & 7.0\% & 1.7\% \\
			
			\midrule
			
			Plain + ALP & 71.5\%  & \textbf{16.3\%} & 23.6\% & \textbf{11.7\%} & \textbf{10.7\%} \\
			
			50\% AT + ALP & 70.4\% & 14.9\% & 21.8\% & 11.5\% & 10.5\% \\
			
			50\% AT & 73.8\% & 11.4\% & 18.6\% & 8.0\% & 7.3\% \\
			
			100\% AT + ALP & 65.7\% & 11.7\% & 19.0\% & 8.1\% & 6.4\% \\
			
			100\% AT & 65.7\% & 11.8\% & 16.0\% & 7.6\% & 6.7\% \\
			\bottomrule
		\end{tabular}%
	\vspace{0.1cm}
	\caption{Clean and adversarial accuracy against different attacks on CIFAR-10. Adversarial accuracy is evaluated against the PGD attack using $\epsilon = 16$ with different number of iterations $n$, step sizes $\epsilon_i$, and restarts $r$. AT denotes adversarial training. CLP and LSQ models are trained with $\lambda=0.25$ and with $\mathcal{N}(0, 0.06)$ noise augmentation. All ALP models were trained with $\lambda=0.5$.}
	\label{table:cifar10}
\end{table}

\subsection{Results on MNIST}

The results of our evaluation on MNIST are given in Table \ref{table:mnist}.  We find that when performing only a single restart of the PGD attack with the default settings, the model trained with LSQ provides an adversarial accuracy of $70.6\%$. However, as can be seen in Figure \ref{fig:grid_search_clp}, the default PGD settings on MNIST ($\epsilon_i = 2.55$, $n=40$) are suboptimal compared to having a larger step size and more iterations. As a result, by increasing the step size as well as the number of iterations and restarts, we can significantly reduce the adversarial accuracy of the LSQ model to $5.0\%$. Following the same approach, we can reduce the adversarial accuracy for the model trained with CLP from $62.4\%$ to $4.1\%$.  We could not achieve a similar reduction in accuracy by using the SPSA attack with $r=1$.

For the adversarially trained models, the situation is different. Even our strongest attack could not reduce the adversarial accuracies of the models combining adversarial training with ALP below $89.9\%$ and $85.7\%$. Further, the ALP model which was trained on clean samples only, achieves a comparable adversarial accuracy of $88.9\%$ against our strongest attack, giving an improvement of $1.7\%$ over the models trained using adversarial training only.

\subsection{Results on CIFAR-10}

Results on CIFAR-10 can be found in Table \ref{table:cifar10}. Again, we find that both CLP and LSQ do not give actual robustness as the accuracy of the models trained using either CLP or LSQ can be reduced to $0.0\%$ and $1.7\%$, respectively. This clearly shows that the robustness of $27.0\%$ of the LSQ model against the baseline PGD attack is misleading. On the other hand, we find that ALP can lead to some robustness even against our strongest PGD attack and outperforms adversarial training by $3.4\%$,~which is in agreement with our findings on MNIST. Note that also for CIFAR-10 we can not achieve a similar drop in accuracy by simply using the SPSA attack.

\subsection{Results on Tiny ImageNet}

\begin{table}[t]
	\centering
	
		\begin{tabular}{lrrrr}
			\toprule
			
			& & \multicolumn{ 3}{c}{\textbf{Adversarial accuracy}~ ($L_{\infty}$, $\epsilon = 16.0$)} \\
			
			\cmidrule(r){3-5}
			
			\multirow{ 3}{*}{\textbf{Model}} & \multirow{ 3}{*}{\textbf{Accuracy}} & \multicolumn{3}{c}{\textbf{PGD attack}} \\
			\cmidrule(r){3-5}
			
			& & \quad $\epsilon_{i} = 2.0$  & \quad $\epsilon_{i} = 4.0$ & $\epsilon_{i} = 4.0$ \\
			& & $n=10$ & $n=400$ & $n=400$ \\
			& & $r=1$ & $r=1$ & $r=100$ \\
			
			\midrule
			
			Plain & 53.0\% & 3.9\% & 0.4\% & 0.4\% \\
			CLP & 48.5\% & 12.2\% & 1.7\% & 0.7\% \\
			LSQ & 49.4\% & 12.8\% & 1.3\% & 0.8\% \\
			
			\midrule
			
			Plain + ALP LL & 53.5\% & 17.2\% & 2.0\% & 0.8\% \\
			Fine-tuned Plain + ALP LL & \textbf{72.0}\% & \textbf{31.8\%} & 10.0\% & 3.6\% \\
			
			\midrule
			
			50\% AT LL & 46.3\% & 25.1\% & 13.5\% & 9.4\% \\
			50\% AT LL + ALP LL & 45.2\% & 26.3\% & 18.7\% & 13.5\% \\
			100\% AT LL & 41.2\% & 25.5\% & 19.5\% & 16.3\% \\
			100\% AT LL + ALP LL & 37.0\% & 25.4\% & \textbf{19.6\%} & \textbf{16.5\%} \\
			
			\bottomrule
		\end{tabular}%
	\vspace{0.1cm}
	\caption{Clean and adversarial top-1 accuracy against different PGD  attacks using a \textit{random target label} on Tiny ImageNet. Adversarial accuracy is evaluated against the PGD attack using $\epsilon = 16$ with a different number of iterations $n$, step sizes $\epsilon_i$, and restarts $r$. The suffix LL denotes that adversarial examples used for training were crafted with the least-likely target. CLP and LSQ models are trained with $\lambda=0.25$ and $\lambda=0.05$ respectively, and augmented by $\mathcal{N}(0, 0.06)$ noise. All ALP models were trained with $\lambda=0.5$.}
	\label{table:tiny_imagenet}
\end{table}

The results of our experiments on Tiny ImageNet are given in Table \ref{table:tiny_imagenet}. Note that we use targeted PGD attacks for crafting adversarial examples and report adversarial accuracy instead of success rate in order to be consistent with \citet{kannan2018adversarial}. We again find that both CLP and LSQ models do not provide actual robustness. Next, we analyze the model provided by \citet{kannan2018adversarial} ``Fine-tuned Plain + ALP LL ($\lambda = 0.5$)'', which was fine-tuned from a model trained on full ImageNet. Our results show that we can reduce the adversarial accuracy from $31.8\%$ to $3.6\%$. This suggests that this model does not provide state-of-the-art robustness against white-box PGD attacks. However, when combined with training only on targeted adversarial examples, ALP marginally improves the adversarial accuracy over plain targeted adversarial training by $0.2\%$ while sacrificing $4.2\%$ of clean accuracy. In additional experiments (Tables \ref{table:tiny_imagenet_full_rnd} and \ref{table:tiny_imagenet_full_ll} in the Appendix) we confirm the hypothesis of \citet{engstrom2018evaluating} that adversarial training using an \textit{untargeted} PGD attack leads to improved adversarial accuracy. Note that \citet{engstrom2018evaluating} can reduce the adversarial accuracy of a pre-trained ALP model trained on full ImageNet to $0.6\%$. In contrast to that, we consider a different model released by \citet{kannan2018adversarial} that was fine-tuned on Tiny ImageNet. This explains the difference in adversarial accuracy reported in Table \ref{table:tiny_imagenet}.

\section{Conclusions}

\begin{figure}[t] 
	\centering
	\resizebox{.95\columnwidth}{!}{%
		\hspace{-.5cm}
		\centering
		\begin{subfigure}[t]{0.225\textwidth}		
			\centering
			\includegraphics[width=1.1\textwidth]{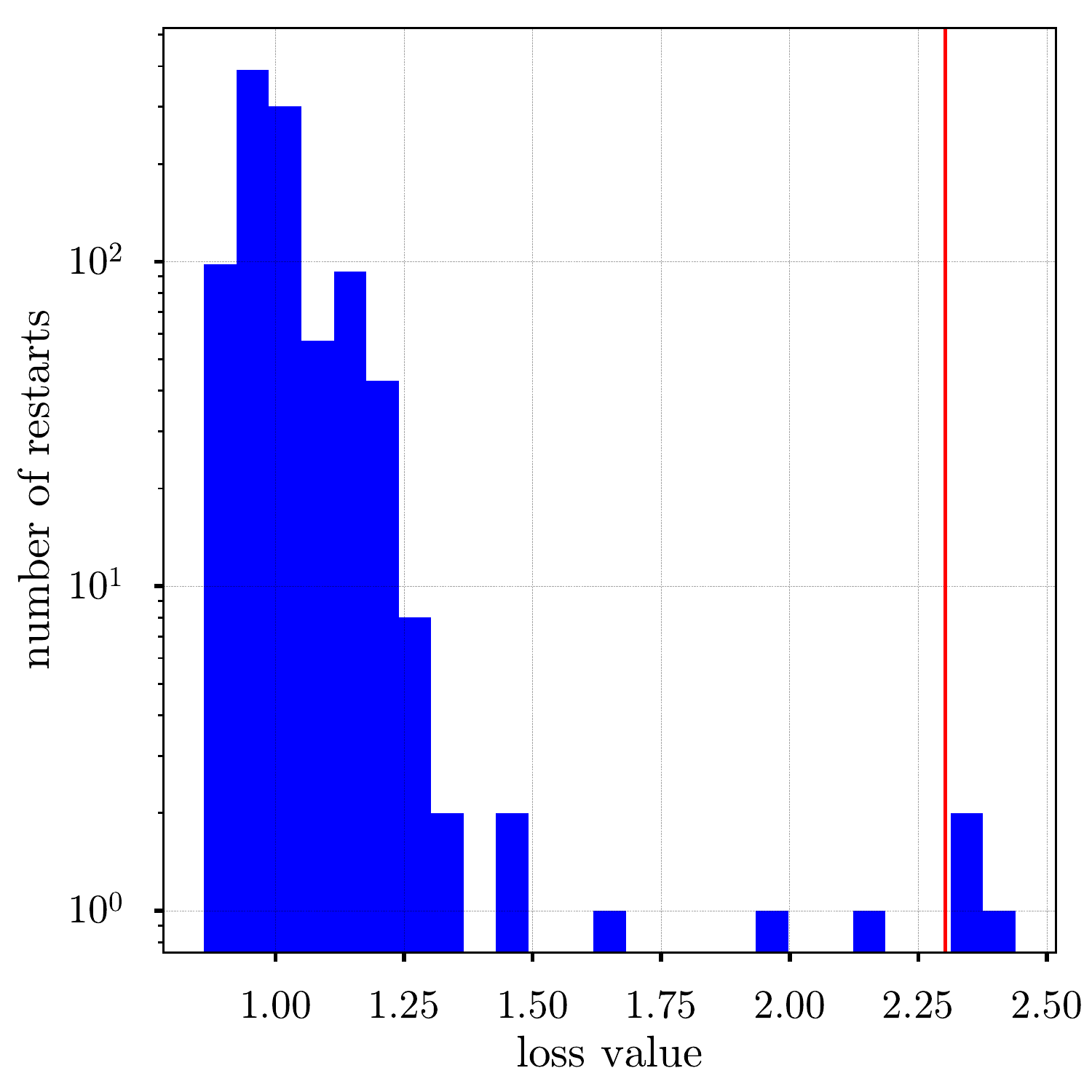}
			\caption{\centering Many restarts are needed}
		\end{subfigure}%
		\quad ~
		\begin{subfigure}[t]{0.225\textwidth}		
			\centering
			\includegraphics[width=1.1\textwidth]{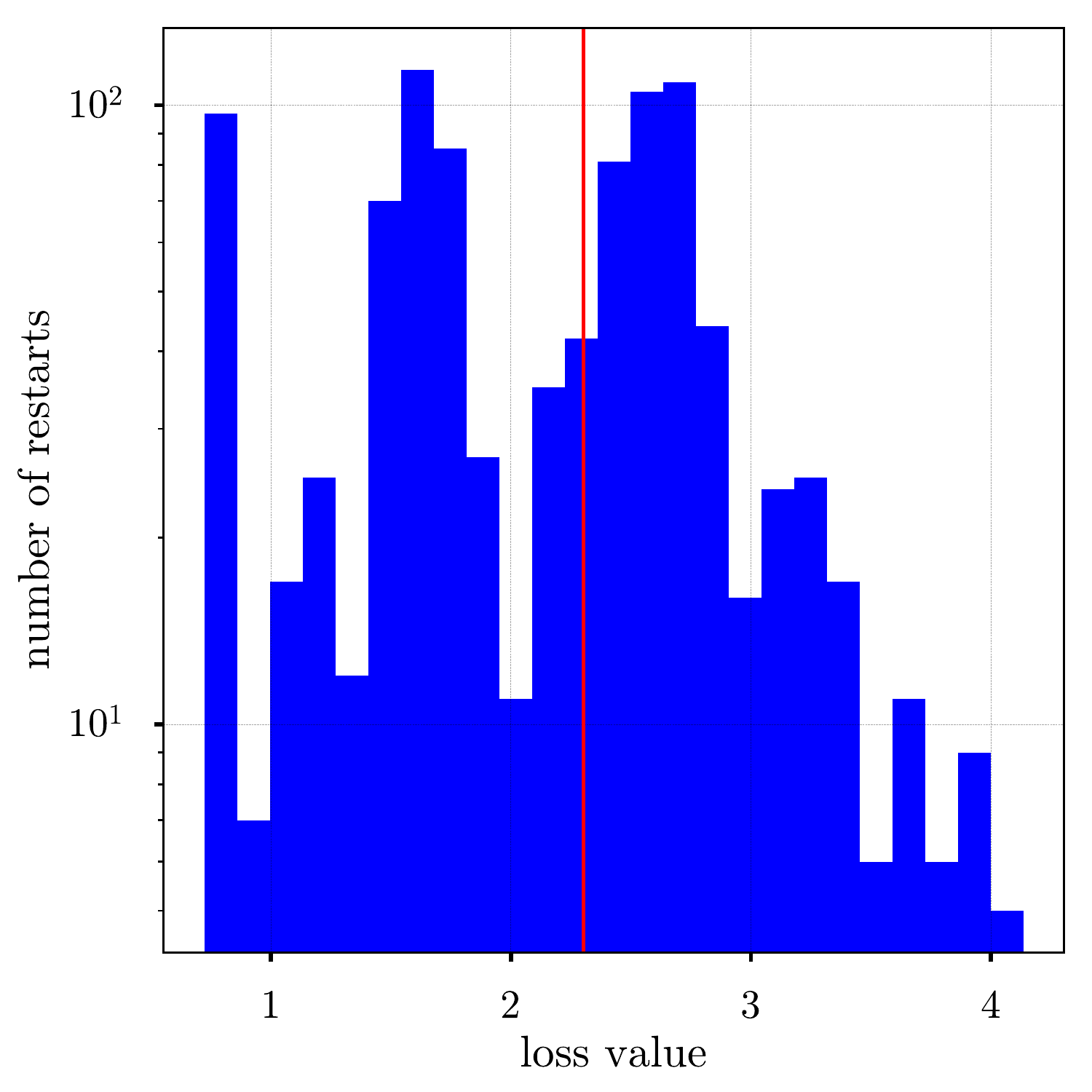}
			\caption{\centering Depends on the restart}
		\end{subfigure}%
		\quad ~
		\begin{subfigure}[t]{0.225\textwidth}		
			\centering
			\includegraphics[width=1.1\textwidth]{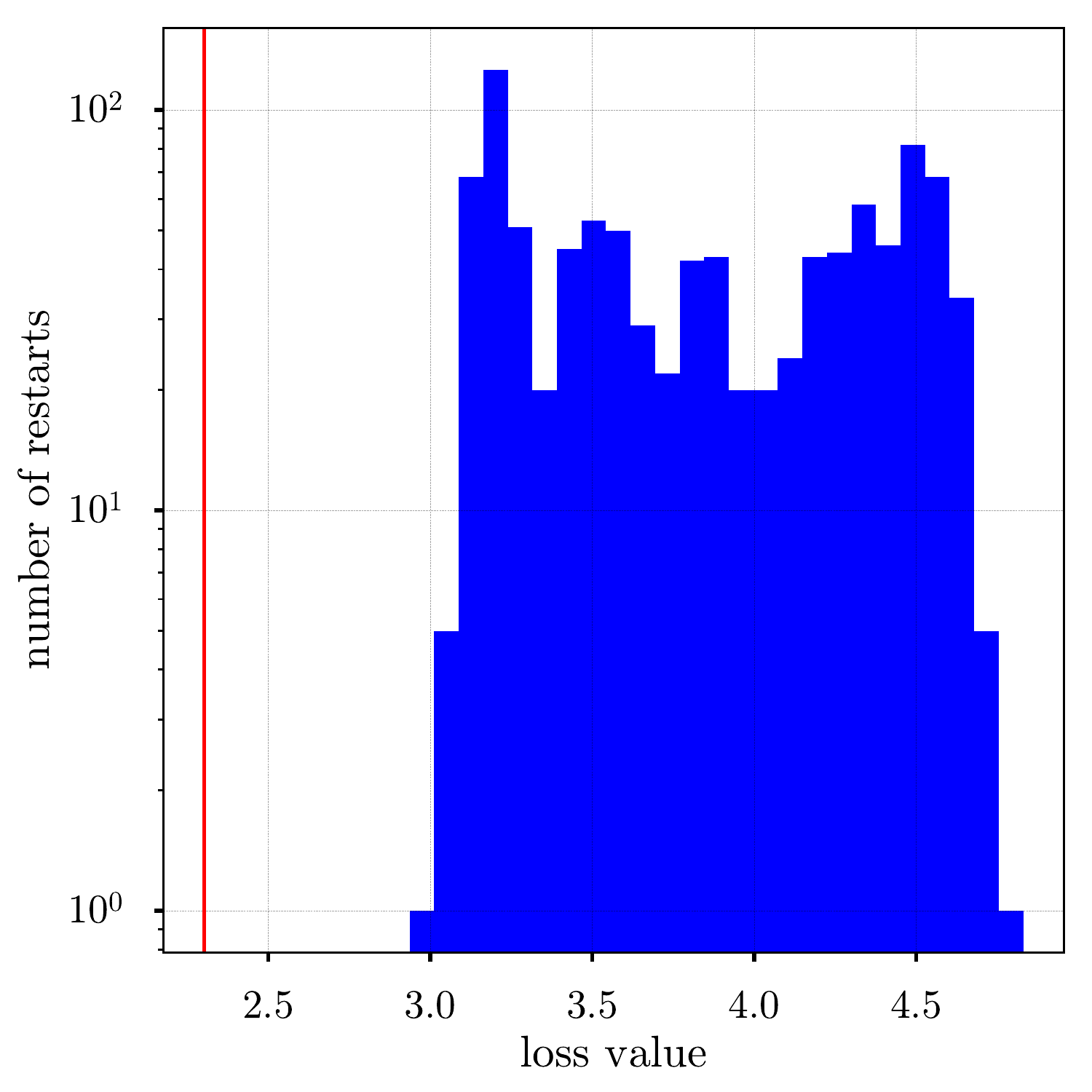}
			\caption{\centering Adv. examples are easily found}
		\end{subfigure}%
		\quad ~
		\begin{subfigure}[t]{0.225\textwidth}		
			\centering
			\includegraphics[width=1.1\textwidth]{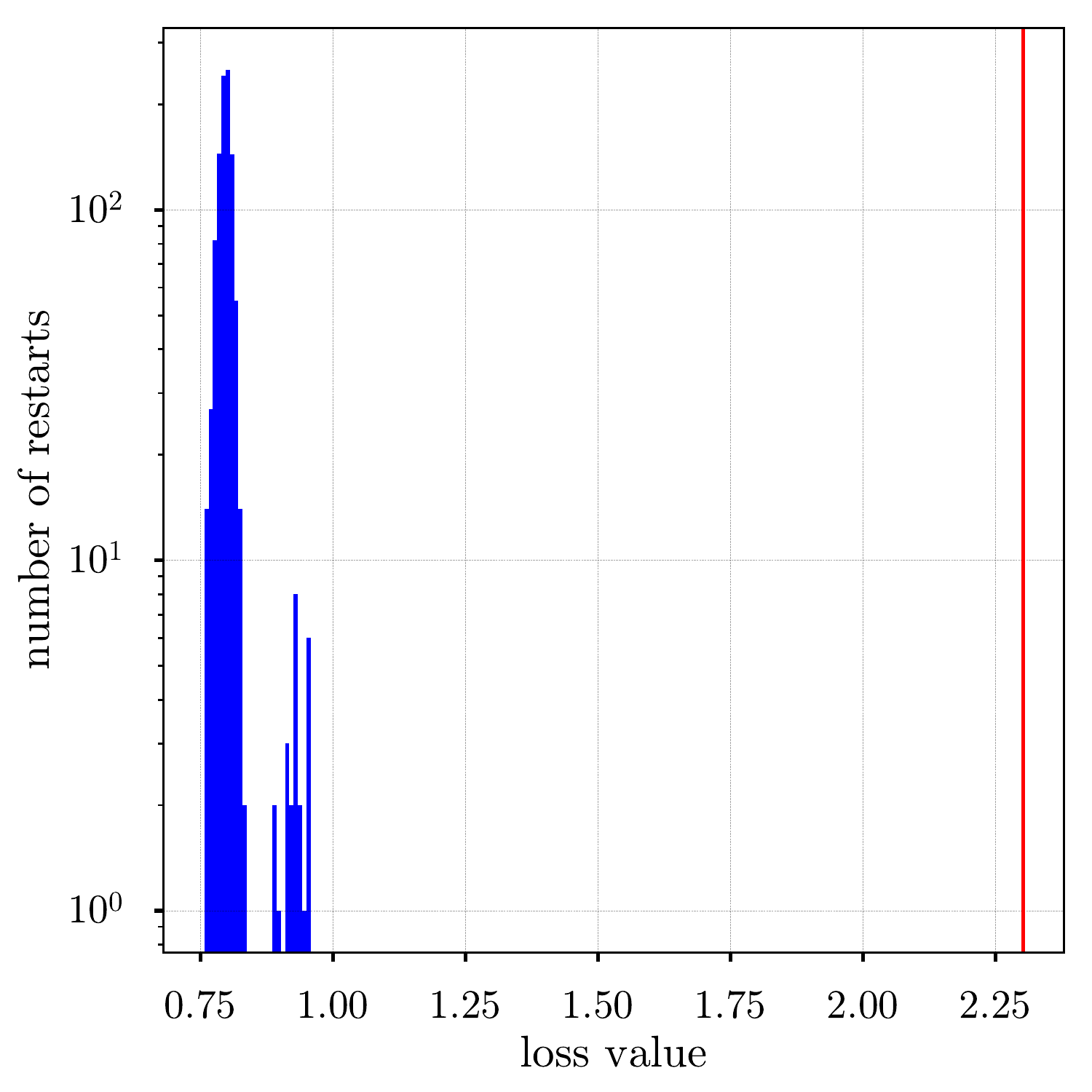}
			\caption{\centering No adv. examples found}
		\end{subfigure}%
	}
	
	\caption{Histograms of the loss values for a single point for $10000$ random restarts of the PGD attack for CLP model trained on MNIST. We show 4 typical cases, which illustrate that there exist points for which the loss can be successfully maximized only with a good starting point. The vertical red line denotes the loss value of $- \ln(0.1)$, which guarantees that for this and higher values of the loss an adversarial example can be found. More histograms can be found in Figure \ref{fig:histograms_appendix} in the Appendix.}
	\label{fig:histograms}
\end{figure}

We perform an empirical evaluation investigating the robustness of logit pairing methods introduced by \citet{kannan2018adversarial}. We find that both CLP and LSQ deteriorate the input space loss surface and make crafting adversarial examples with gradient-based attacks difficult, without providing actual robustness. This suggests that the current practice of evaluating against the PGD attack with default settings can be misleading. Therefore, one should consider performing an exhaustive grid search over the PGD attack parameters in addition to performing many random restarts of PGD, which helps to find adversarial examples in such cases (Figures~\ref{fig:grid_search_clp}, \ref{fig:histograms}). Finally, we show that the ALP models of \citet{kannan2018adversarial} do not improve drastically over adversarial training alone.

\paragraph{Acknowledgments.} We would like to thank Michael Hedderich, Francesco Croce, and Dave Howcroft for their helpful feedback on this paper. Furthermore, we thank the reviewers for their valuable comments. Marius Mosbach acknowledges partial support by the German Research Foundation (DFG) as part of SFB 1102.
 
\bibliography{references}

\newpage
\section*{Appendix A}

\section*{Additional results and visualizations}

\begin{figure}[h] 
	\centering
	\resizebox{.95\columnwidth}{!}{%
	\centering
	\includegraphics[width=0.9\textwidth]{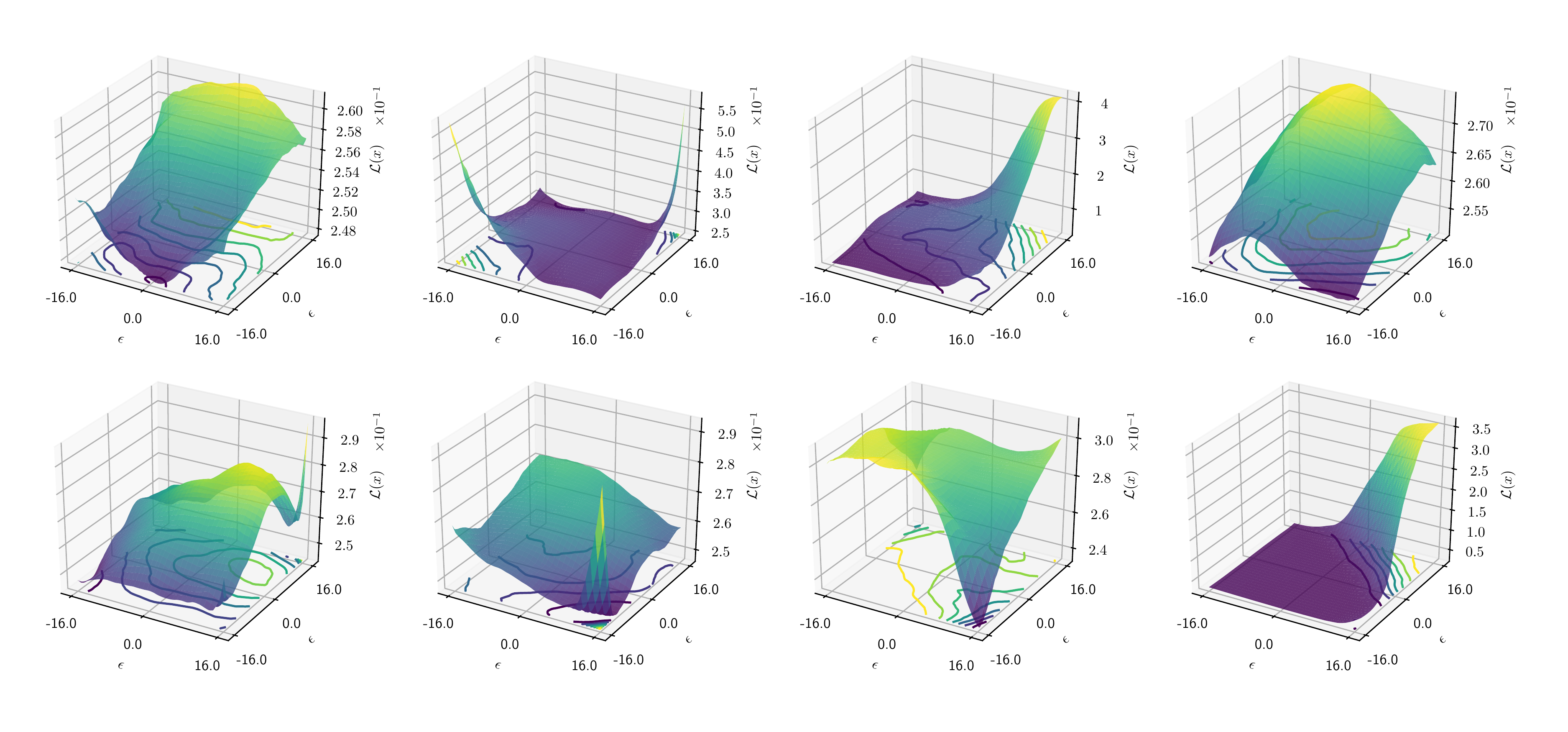}%
	}
	\caption{Input loss surfaces of LSQ model on CIFAR10 in a random subspace with $\epsilon=16.0$ for the first eight test examples. For some points, the loss surface contains local maxima and thus may pose a problem for gradient-based attacks. This is in line with our quantitative results in Table \ref{table:cifar10}, showing that this model does not provide actual robustness, and a successful gradient-based attack must use many random restarts of the PGD attack.}
	\label{fig:cifar10_clp_loss_surfaces_appendix}
\end{figure}

\begin{figure}[h] 
	\centering
	\resizebox{.95\columnwidth}{!}{%
	\centering
	\includegraphics[width=0.9\textwidth]{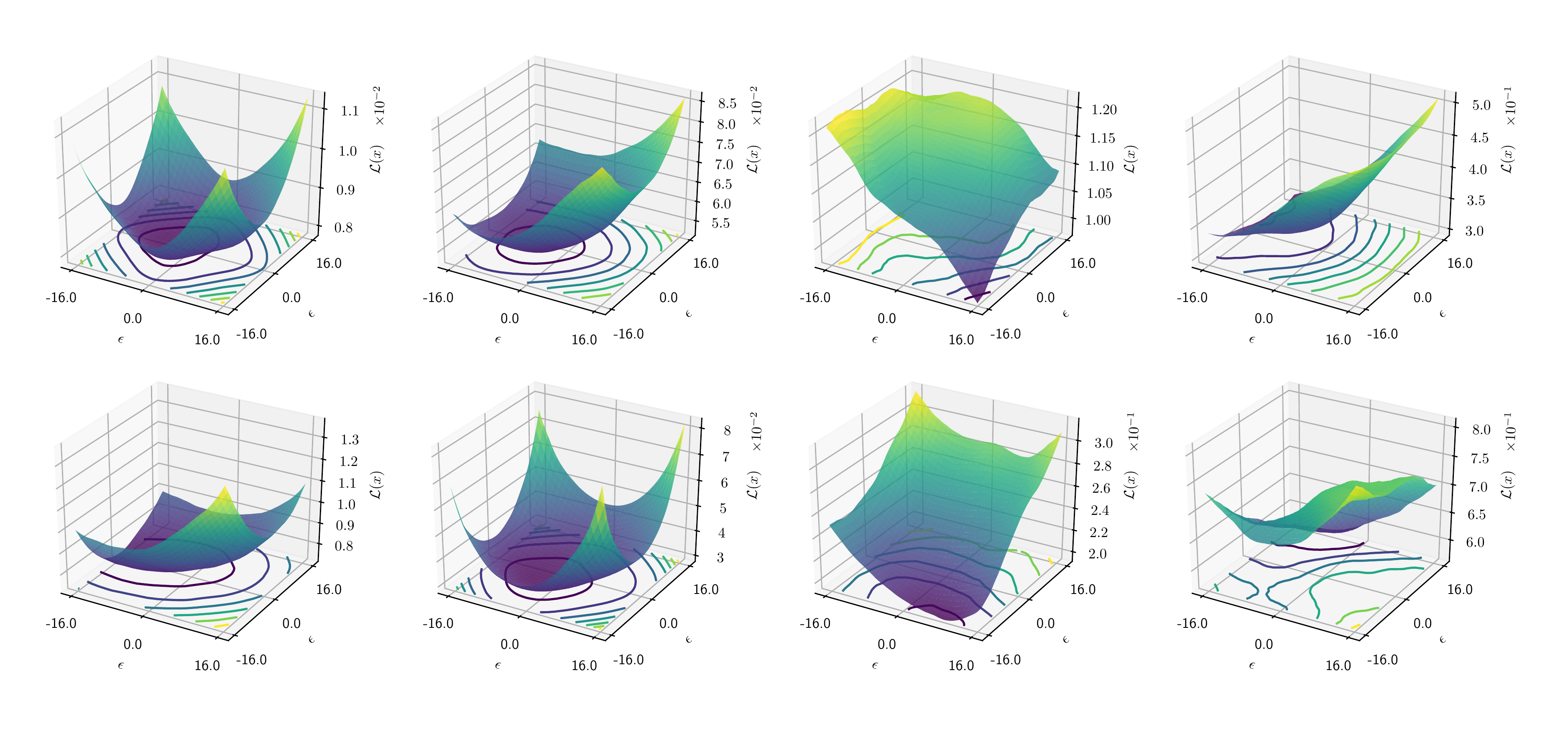}%
	}
	\caption{Input loss surfaces of Plain + ALP model on CIFAR-10 in a random subspace with $\epsilon=16.0$ for the first eight test examples. The loss surface may be suitable for gradient descent. This is in line with our quantitative results in Table \ref{table:cifar10}, showing that there is only a small benefit in using random restarts of the PGD attack.}
	\label{fig:cifar10_alp_loss_surfaces_appendix}
\end{figure}

\begin{figure}[p] 
	\centering
	\resizebox{.95\columnwidth}{!}{%
	\centering
	\includegraphics[width=0.9\textwidth]{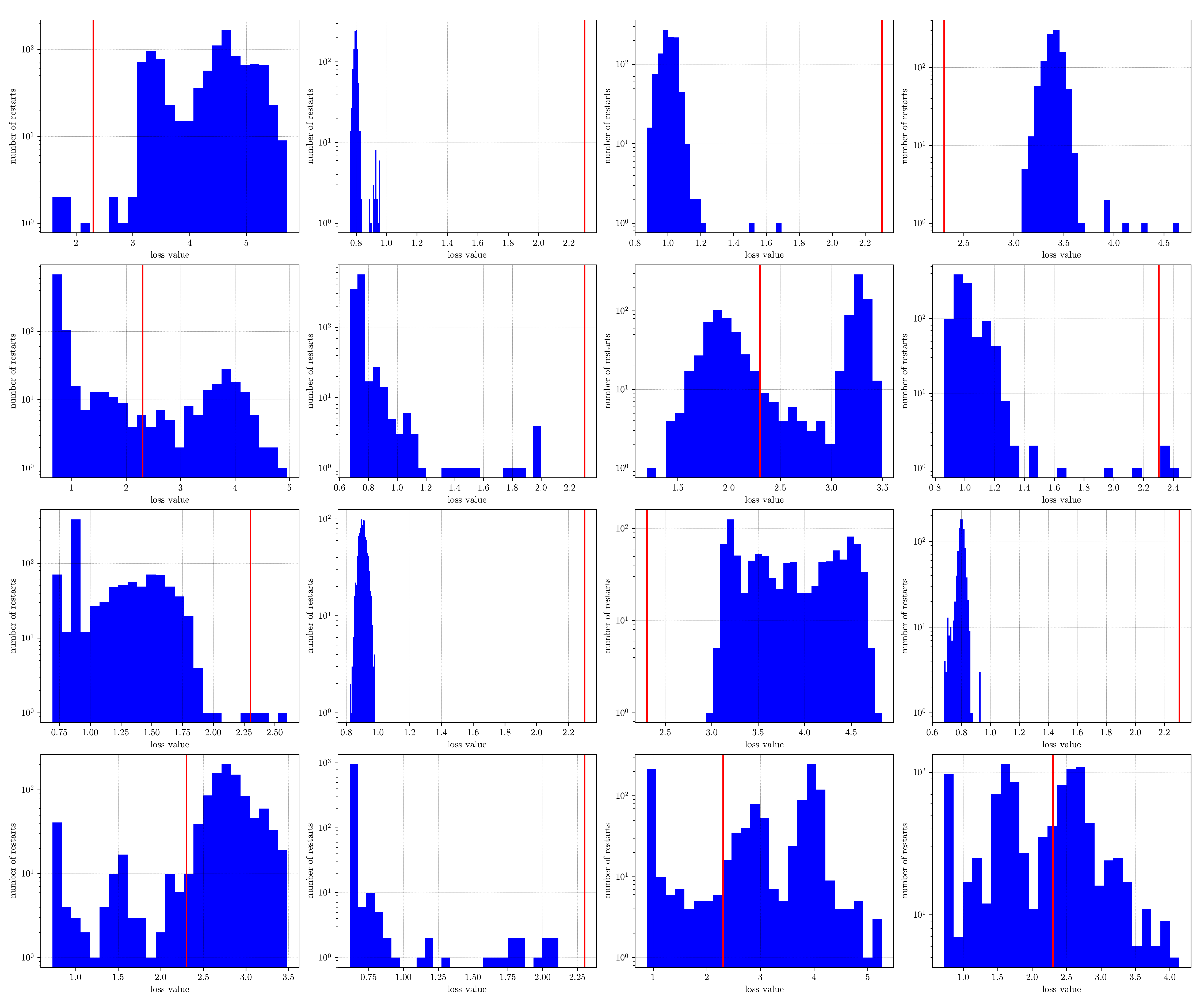}%
	}
	\caption{Histograms of the loss values across $10000$ random restarts of the PGD attack on CLP model trained on MNIST for the first 16 test examples. The vertical red line denotes the loss value of~$- \ln(0.1)$, which guarantees that for this and higher values of the loss an adversarial example can be found.}
	\label{fig:histograms_appendix}
\end{figure}

\begin{table}[h]
	\centering
		\begin{tabular}{lrrrr}
			\toprule
			
			& & \multicolumn{ 3}{c}{\textbf{Adversarial accuracy}~ ($L_{\infty}$, $\epsilon = 16.0$)} \\
			
			\cmidrule(r){3-5}
			
			\multirow{ 3}{*}{\textbf{Model}} & \multirow{ 3}{*}{\textbf{Accuracy}} & \multicolumn{3}{c}{\textbf{PGD attack, random target}} \\
			\cmidrule(r){3-5}
			
			& & \quad $\epsilon_{i} = 2.0$  & \quad $\epsilon_{i} = 4.0$ & $\epsilon_{i} = 4.0$ \\
			& & $n=10$ & $n=400$ & $n=400$ \\
			& & $r=1$ & $r=1$ & $r=100$ \\
			
			\midrule
			
			Plain & 53.0\% & 3.9\% & 0.4\% & 0.4\% \\
			CLP & 48.5\% & 12.2\% & 1.7\% & 0.7\% \\
			LSQ & 49.4\% & 12.8\% & 1.3\% & 0.8\% \\
			\midrule
			Plain + ALP & 53.3\% & 17.4\% & 3.1\% & 1.4\% \\
			Plain + ALP LL & 51.5\% & 12.2\% & 0.9\% & 0.3\% \\
			Fine-tuned Plain + ALP LL & \textbf{72.0}\% & \textbf{31.8\%} & 10.0\% & 3.6\% \\
			
			\midrule
			
			50\% AT & 54.7\% & 23.1\% & 21.5\% & 17.9\% \\
			50\% AT + ALP & 37.4\% & 26.1\% & \textbf{23.2\%} & \textbf{20.1\%} \\
			50\% AT LL & 46.3\% & 25.1\% & 13.5\% & 9.4\% \\
			50\% AT LL + ALP LL & 45.2\% & 26.3\% & 18.7\% & 13.5\% \\
			100\% AT & 19.6\% & 15.4\% & 15.0\% & 12.5\% \\
			100\% AT LL & 41.2\% & 25.5\% & 19.5\% & 16.3\% \\
			100\% AT LL + ALP LL & 37.0\% & 25.4\% & 19.6\% & 16.5\% \\
			
			\bottomrule
		\end{tabular}%
	\vspace{0.1cm}
	\caption{Clean and adversarial top-1 accuracy against different PGD  attacks using a \textit{random target label} on Tiny ImageNet. The suffix LL denotes that adversarial examples used for training were crafted with the least-likely target. If omitted, adversarial examples were crafted using an untargeted attack. CLP and LSQ models are trained with $\lambda=0.25$ and $\lambda=0.05$ respectively, and augmented by $\mathcal{N}(0, 0.06)$ noise. All ALP models were trained with $\lambda=0.5$.}
	\label{table:tiny_imagenet_full_rnd}
\end{table}

\begin{table}[h]
	\centering
		\begin{tabular}{lrrrr}
			\toprule
			
			& & \multicolumn{ 3}{c}{\textbf{Adversarial accuracy}~ ($L_{\infty}$, $\epsilon = 16.0$)} \\
			
			\cmidrule(r){3-5}
			
			\multirow{ 3}{*}{\textbf{Model}} & \multirow{ 3}{*}{\textbf{Accuracy}} & \multicolumn{3}{c}{\textbf{PGD attack, least-likely target}} \\
			\cmidrule(r){3-5}
			
			& & \quad $\epsilon_{i} = 2.0$  & \quad $\epsilon_{i} = 4.0$ & $\epsilon_{i} = 4.0$ \\
			& & $n=10$ & $n=400$ & $n=400$ \\
			& & $r=1$ & $r=1$ & $r=100$ \\
			
			\midrule
			
			Plain & 53.0\% & 5.6\% & 0.1\% & 0.0\% \\
			CLP & 48.5\% & 11.6\% & 0.9\% & 0.0\% \\
			LSQ & 49.4\% & 11.5\% & 0.3\% & 0.0\% \\
			Plain + ALP & 53.3\% & 23.7\% & 6.2\% & 1.3\% \\
			\midrule
			Plain + ALP LL & 51.5\% & 17.0\% & 1.9\% & 0.0\% \\
			Fine-tuned Plain + ALP LL & \textbf{72.0\%} & \textbf{38.3\%} & 14.8\% & 4.2\% \\
			\midrule
			50\% AT & 54.7\% & 19.6\% & 19.2\% & 15.5\% \\
			50\% AT + ALP & 37.4\% & 29.4\% & 27.1\% & \textbf{22.2\%} \\
			50\% AT LL & 46.3\% & 29.8\% & 22.6\% & 13.3\% \\
			50\% AT LL + ALP LL & 45.2\% & 28.7\% & 25.6\% & 17.4\% \\
			100\% AT & 19.6\% & 16.0\% & 15.2\% & 13.2\% \\
			100\% AT LL & 41.2\% & 28.9\% & \textbf{27.8\%} & 21.8\% \\
			100\% AT LL + ALP LL & 37.0\% & 27.6\% & 25.7\% & 19.7\% \\
			
			\bottomrule
		\end{tabular}%
	\vspace{0.1cm}
	\caption{Clean and adversarial top-1 accuracy against different PGD  attacks using the \textit{least-likely target label} on Tiny ImageNet. The suffix LL denotes that adversarial examples used for training were crafted with the least-likely target. If omitted, adversarial examples were crafted using an untargeted attack. CLP and LSQ models are trained with $\lambda=0.25$ and $\lambda=0.05$ respectively, and augmented by $\mathcal{N}(0, 0.06)$ noise. All ALP models were trained with $\lambda=0.5$.}
	\label{table:tiny_imagenet_full_ll}
\end{table}

\end{document}